\documentclass[conference]{IEEEtran}
\usepackage{cite}
\usepackage{amsmath,amssymb,amsfonts}
\usepackage{algorithmic}
\usepackage{graphicx}
\usepackage{textcomp}
\usepackage{xcolor}
\def\BibTeX{{\rm B\kern-.05em{\sc i\kern-.025em b}\kern-.08em
    T\kern-.1667em\lower.7ex\hbox{E}\kern-.125emX}}
    
\graphicspath{ {./} }

\begin{document}

\title{Estimating Player Completion Rate in Mobile Puzzle Games Using Reinforcement Learning
}

\author{\IEEEauthorblockN{Jeppe Theiss Kristensen}
\IEEEauthorblockA{ 
\textit{IT University of Copenhagen/Tactile Games}\\
Copenhagen, Denmark \\
jetk@itu.dk}
\and
\IEEEauthorblockN{Arturo Valdivia}
\IEEEauthorblockA{
\textit{Tactile Games}\\
Copenhagen, Denmark \\
arturo@tactile.dk}
\and
\IEEEauthorblockN{Paolo Burelli}
\IEEEauthorblockA{
\textit{IT University of Copenhagen/Tactile Games}\\
Copenhagen, Denmark \\
pabu@itu.dk}
}


\maketitle

\begin{abstract}



In this  work we investigate whether it is plausible to use the performance of a reinforcement learning (RL) agent to estimate the difficulty measured as the player completion rate of different levels in the mobile puzzle game Lily's Garden.

For this purpose we train an RL agent and measure the number of moves required to complete a level. This is then compared to the level completion rate of a large sample of real players.

We find that the strongest predictor of player completion rate for a level is the number of moves taken to complete a level of the $\sim$5\% best runs of the agent on a given level. A very interesting observation is that, while in absolute terms, the agent is unable to reach human-level performance across all levels, the differences in terms of behaviour between levels are highly correlated to the differences in human behaviour.
Thus, despite performing sub-par, it is still possible to use the performance of the agent to estimate, and perhaps further model, player metrics.

\end{abstract}


\section{Introduction}

Automatic testing of games has long been one of the objectives of research in game artificial intelligence. 
The ability to use an agent to test new content and mechanics has the potential to dramatically reduce the cost of production of games and improve the game designers' workflow. 

Over the years, autonomous game-playing agents have been developed using different techniques ranging from rule based systems to machine learning methods such as supervised learning or reinforcement learning. 
While the objective in large parts of these agents is to play the game optimally -- \emph{i.e.} to solve the game -- many efforts have been put also into creating agents that behave in a way that resembles as closely as possible the way a human player would play \cite{Hingston2009ABots}.


The purpose of human-like agents is to play a game competently and behave in a way that is indistinguishable from a human player to an outside observer; essentially, being able pass a game version of the Turing test~\cite{Togelius2012}. Agents of this kind are ideal candidates to evaluate game content as the observation of their behaviour in the game can give a more realistic feedback to a game designer.



\begin{figure*}[ht]
    \centering
    \includegraphics[width=0.80\columnwidth]{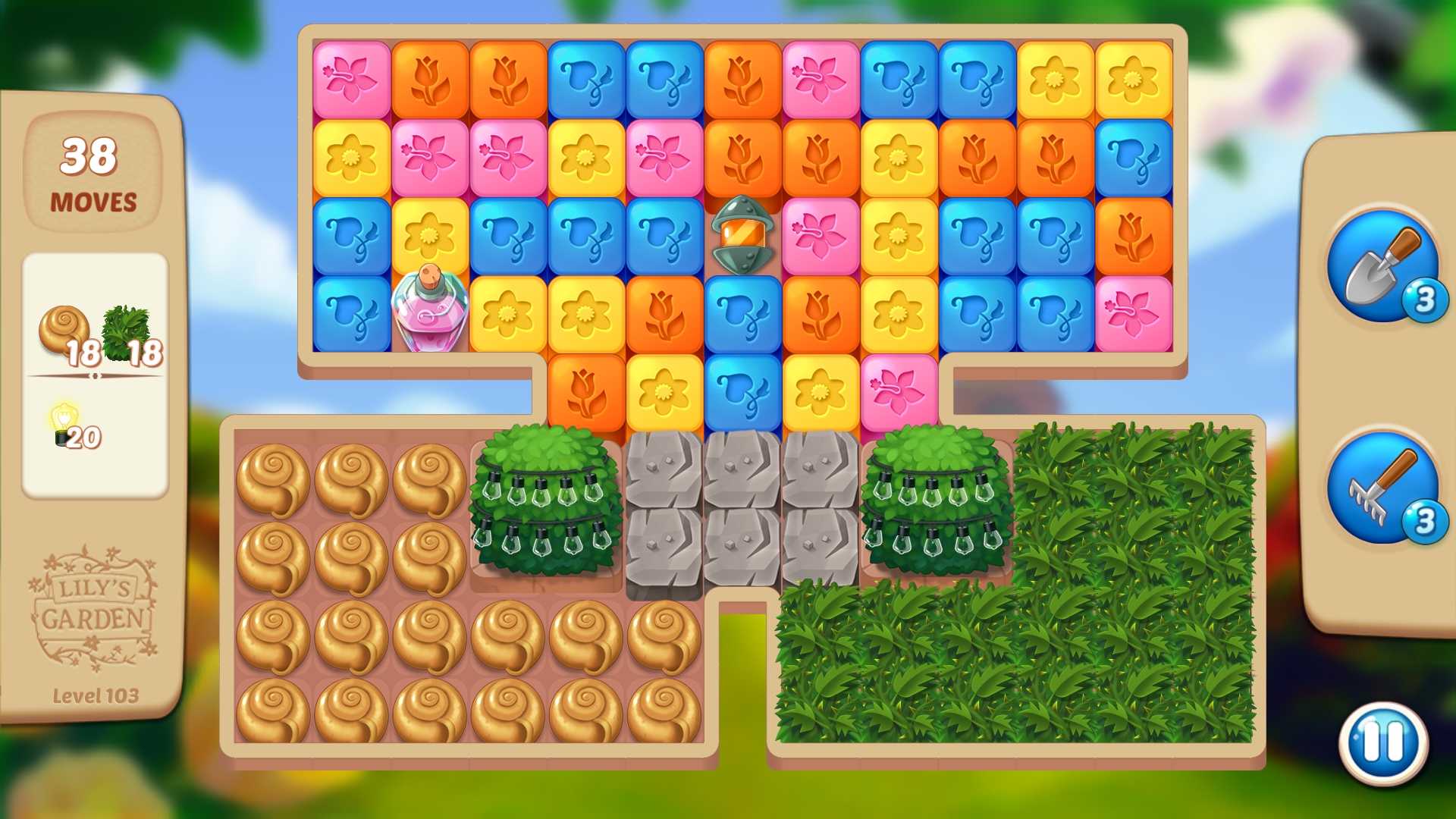}
    \includegraphics[width=0.9\columnwidth]{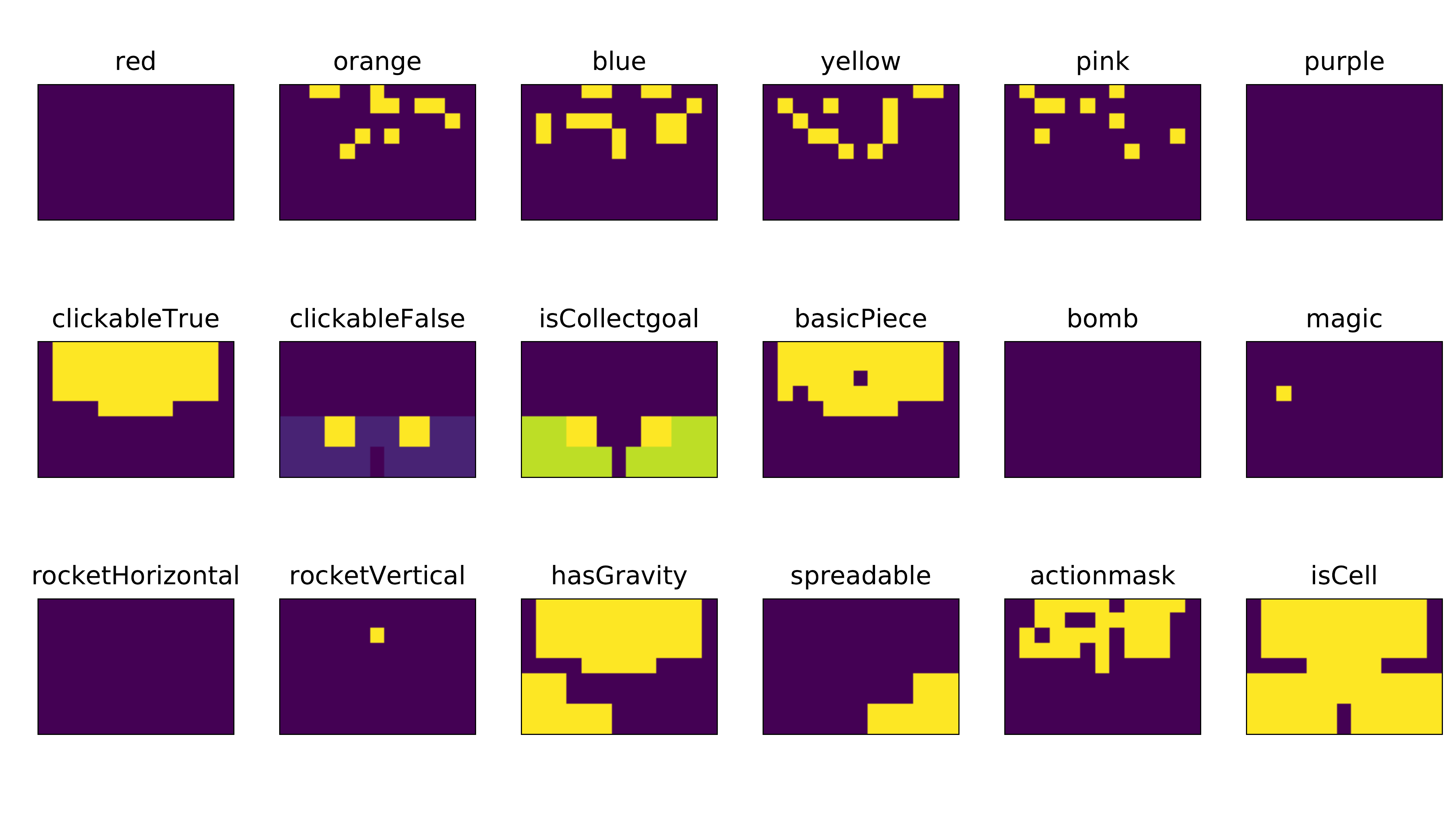}
    \caption{Example of how the game-board  of level 103 looks like \textit{left)} in-game and \textit{right)} how it is represented in our custom environment. The channels are not one-hot encoded but use the hit points of the board piece, hence the different colours in the \textsc{clickableFalse} and \textsc{isCollectgoal} channels.}
    \label{fig:level example}
\end{figure*}



Creating human-like agents for game-play testing has been explored in a number of other works.
Holmgård et al.\cite{Holmgard2018} explore creating such agents using MCTS and evolving node selection criteria in order to generate procedural player personas. Mugrai et al.~\cite{Mugrai2019AutomatedGames} developed this method further and tested it on a puzzle game comparing its scores to the ones of a small number of human-players that participated in their experiment.

Shin et al.~\cite{Shin2020PlaytestingLearning}, in order to try to mirror human players' behaviour, train an RL agent to learn which of 5 predefined human-like strategies to pick before picking a valid action matching the preferred strategy.

Lastly, actual player play-traces can also be used to learn how actual players play, which was demonstrated by Gudmundsson et al.~\cite{Gudmundsson2017}, where a convolutional neural network action selection policy is learned from the play-traces.
However, one issue with using playtraces is that these data are not always available, for example for a newly released game with little or no player data, or technical limitations such as cost of storage or tracking issues.
Thus, an agent trained using reinforcement learning may be a more viable solution.

The initial results on using RL agents for playing Lily's Garden in Kristensen et al.~\cite{Kristensen2020StrategiesGames} show that it is indeed possible to use an RL agent in this setting.
In this research work we expand upon this line of investigation by including more levels and consider the next step for estimating player level completion rate using agent performance.
We investigate how to train and use autonomous agents for estimating the player completion rate of a number of levels in the game Lily's Garden by Tactile Games\footnote{https://tactilegames.com/lilys-garden/}.
For this purpose, we developed a set of Proximal Policy Optimisation (PPO) based reinforcement learning agents~\cite{Schulman2017} and evaluate how the number of steps taken by the agent for completing the levels relate to the behaviour of a sample of $\sim$900,000 players.

\section{Method}

Before outlining the experimental approach, in this section we first present the RL setup that we use for the experiments followed by a description of the evaluation method.

\subsection{Reinforcement Learning Setup}

\begin{figure}[ht]
    \centering
    \includegraphics[width=0.85\columnwidth]{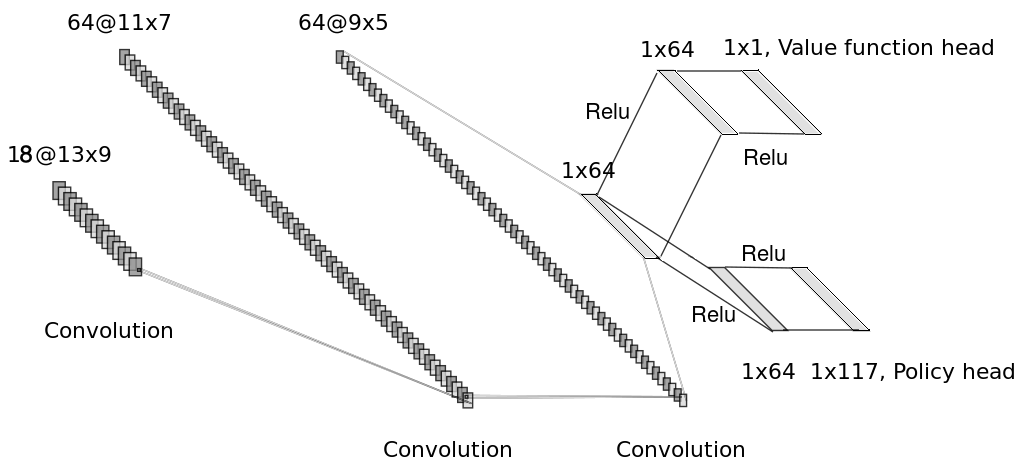}
    \caption{Agent CNN policy setup. Each convolution uses a 2x2 kernel.}
    \label{fig:neural network setup}
\end{figure}

To serve as a test bed for our agent, we use a custom environment of Lily's Garden, detailed in previous work~\cite{Kristensen2020StrategiesGames}.
The game board representation is a (13$\times$9$\times$\textit{m}) array, where the different board piece attributes are encoded in the \textit{m} channels (see Fig. \ref{fig:level example}).



Our RL agent implementation is also based on our previous work in \cite{Kristensen2020StrategiesGames} and follows the on-policy implementation of PPO available in OpenAI Baselines~\cite{openaiGYM} and stable-baselines~\cite{stable-baselines} where multiple agents can collect state-action-reward observations simultaneously.
The architecture of the policy and value networks is shown in Fig. \ref{fig:neural network setup}: three convolutional layers for feature learning are in common between the two networks, with separate value and policy heads.

In all the experiments, we use an entropy coefficient of $0.01$, a learning rate of $1e-4$ and run 8 agents simultaneously.
The other hyper-parameters are set to $n_{\textup{minibatches}} = 64$ (default 4), $n_{\textup{steps}} = 256$ (default 128) and otherwise default values.

Additionally, based on our previous research~\cite{Kristensen2020StrategiesGames}, we use three strategies to improve the stability of the training:
\begin{itemize}
    \item Colour shuffling, where the colour channels are randomly permuted during the training. This is done to help the agent to generalise patterns regardless of the colours.
    \item Resetting after 100 total steps, to prevent the agent from getting stuck on an invalid move and filling the training data buffer with useless observations.
    \item Adding an action mask to the observations, which serves as a partial forward model leading to  faster initial training.
\end{itemize}

\subsection{Estimating player success rate}
\label{subsec:method:difficulty evaluation}

When level designers wish to evaluate a level, the player completion rate is often then used as a proxy for difficulty.
Since players can only complete the level once before progressing, the completion rate can be directly calculated as the number of level completions over total attempts across all users.
It is important to note, though, that this is not necessarily the inherent difficulty of a level and may change in time depending on which cohorts of players that have reached a given point.

\begin{figure*}[th]
    \centering
    \includegraphics[width=1.65\columnwidth]{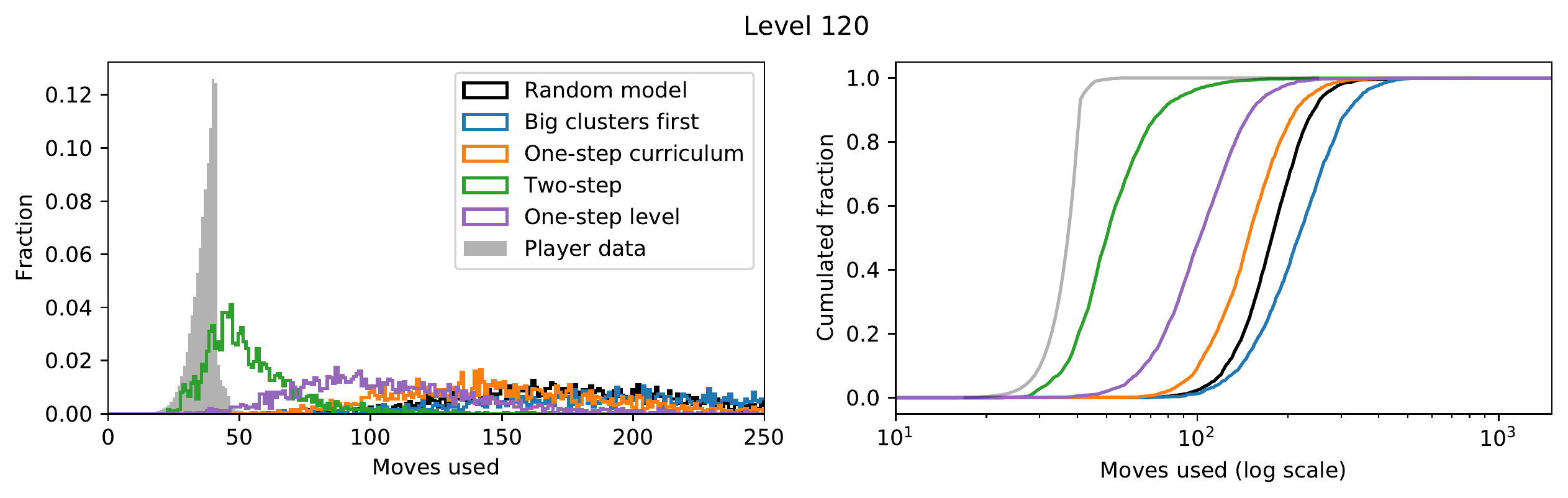}
    \caption{\textit{Left)} Distribution of number of moves required for each model to complete level 120 compared to actual player data. The sharp drop-off in the player distribution of because of the level move limit. Additional steps after the move limit can be purchased using in-game currency. \textit{Right)} Cumulative move distribution, or level completion curves, plotted on a log scale.}
    \label{fig:moves histogram lvl120}
\end{figure*}

In our game environment of Lily's Garden, the agent is able to take an unlimited number of moves per level.
The move distribution is therefore different compared to player data, where there is a sharp cut-off after the move limit, see Fig. \ref{fig:moves histogram lvl120} \textit{left}.
In order compare these two distributions and estimate the player completion rate, we record the max number of moves number of the best \textit{x\%} agent runs and then calculate the Spearman correlation of this number with the player completion rate.
Because the number of moves spent by the agent to complete a level is invariant level move limit is, while the player completion rate is very much determined by this limit, we normalise the agent moves with the level move limit.

\begin{figure}[ht]
    \centering
    \includegraphics[width=0.80\columnwidth]{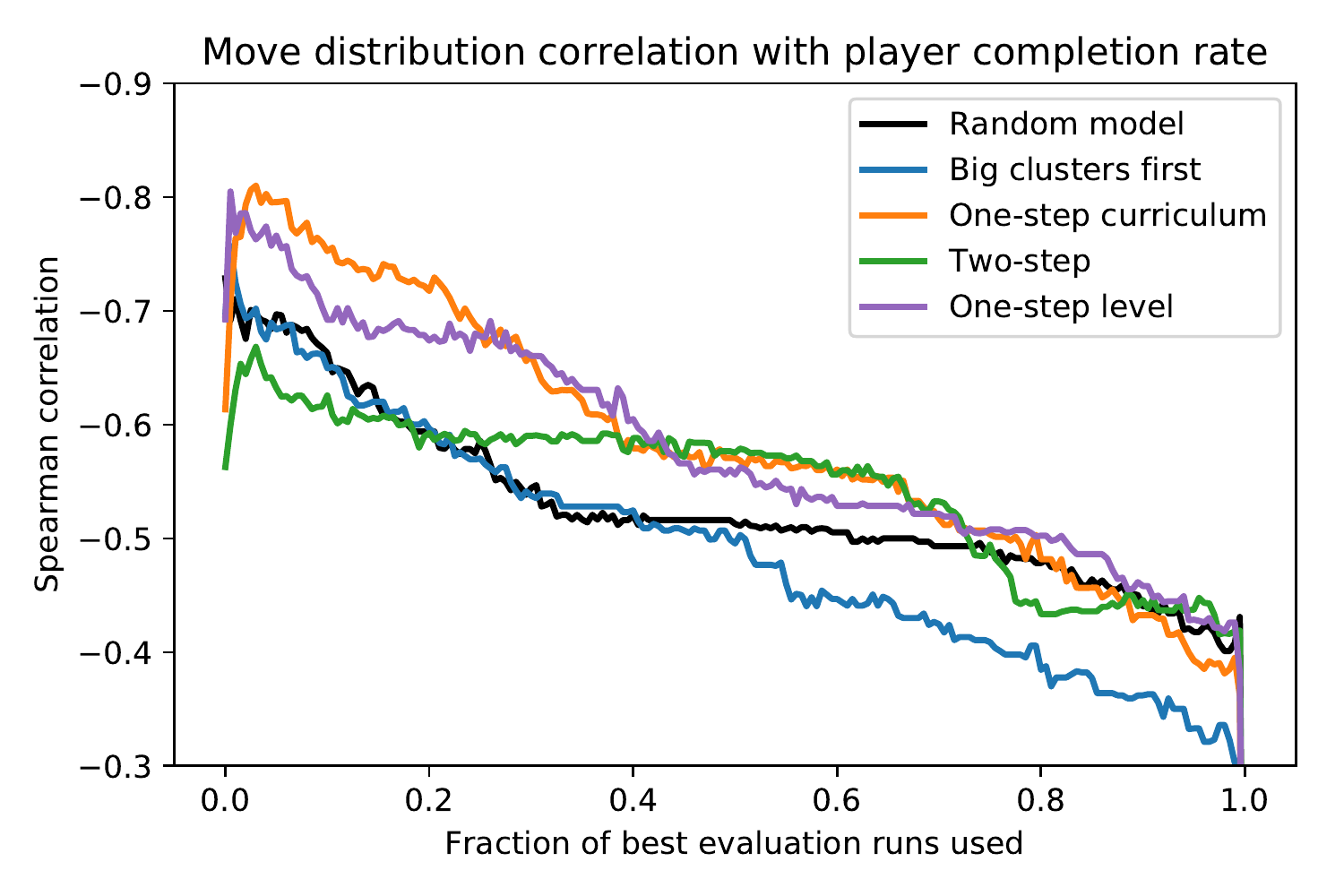}
    \caption{Spearman correlation between player completion rate and normalised moves required to finish each level as a function of fraction of best evaluation runs used.}
    \label{fig:correlation vs completion threshold}
\end{figure}

\section{Experiments}

The objective of this research project is to help determine how an agent can be used in a production setting in a mobile gaming company where not only accuracy but also speed is important. For that purpose, we explore here three scenarios:

\begin{itemize}
    \item One-step training on curriculum
    \item One-step training on the target level
    \item Two-step training, first on a curriculum and then on the target level
\end{itemize}
In each scenario the aforementioned training and in-game behaviour performance are collected and compared to the human behaviour. In addition to that, we benchmark these agents against two baselines:
\begin{itemize}
    \item \emph{Random agent} which picks a random valid action every time it has to make a move.
    \item \emph{Greedy agent} which mimics a play style that prioritises clicking on valid pieces belonging to the biggest clusters.
\end{itemize}

Out the three aforementioned training scenarios, we start with the \emph{One-step training on curriculum} because it offers one desirable feature for production usage: a previously trained agent is used in order to estimate the difficulty of new unseen levels, without requiring any further training.

The training curriculum here should be chosen in such a way that the agent is subsequently able to generalise across new levels and mechanics. Thus, in particular, the curriculum has to include a variety of different types of levels and be representative of different game mechanics (\emph{e.g.}, \emph{rock blockers}, \emph{grass spreading blocks}). Furthermore, for evaluation the agent has to be exposed to a set of equally different unseen levels.

To choose the levels for the curriculum, we note that at the beginning of Lily's Garden, the new mechanics are typically introduced once every ten levels (\emph{i.e.}, on levels 21, 31, 41, ...). The following nine levels at each interval generally mix the new mechanic with previously introduced mechanics. And so for the training, we include the first hundred levels which introduces eight new board pieces in addition to the base mechanics of the game. These levels are uniformly randomly sampled and trained on for at least one epoch over 35M \emph{steps} -- \emph{i.e.}, the equivalent of a human player tapping on the screen. We remark here that from our empirical observations, this number of steps ensures that the learning has plateaued and each level will have been trained on for multiple epochs.

To evaluate the agent performance, we test on the 20 levels following the ones used in training (\emph{i.e.}, levels 101 to 120), which include the previous mechanics plus two new mechanics: \emph{teleporters}, which move board pieces to other parts of the game board, and \emph{containers}, which are 2x2 blockers with 10 hit points.
For the results we also include the levels 13, 23, ..., 93 to test the performance on previous levels which are not tutorial levels.

In the second scenario considered, \emph{One-step training on evaluation level}, the agent is trained every time from scratch directly on the new target level. This is done through 1M steps. 
Two potential drawbacks of this approach are apparent: 
one is that solution may require longer training time before reaching a level of competency when compared to the other scenarios based on curricula. 
Secondly, it may also lead to poor generalisation. 
However, if good accuracy is obtained and the training can still performed in reasonable time, then this approach may still offer a viable solution to used on a production environment.

The final scenario we consider, \emph{Two-step training}, is inspired by how players typically learn: At first the player has some previous general knowledge of how to play the game but no specific knowledge on how to beat certain level or game mechanic. Then, after having played through the level a number of times, the player may finally learn a winning strategy and complete the level. In this setting, the training for the first step is analogous to what was is done in the setting of the \emph{One-step training on curriculum} scenario, while for the second step we proceed as analogously to the \emph{One-step training on evaluation level} scenario.

\section{Results and discussion}


Before answering the question of whether we can correlate the agent behaviour to the players', we first examine which agents that acquire the highest proficiency, as measured by the least amount of average moves spent to complete a level.
A representative example of the level of proficiency the different agents acquire can be seen by considering the move distributions in Fig. \ref{fig:moves histogram lvl120}.
The most proficient agent comes from the two-step training approach, followed by the one-step training on target level and then finally one-step training on a curriculum.

All the training scenarios lead to agents performing better than the random and greedy agents.
Despite the fact that some agents only perform slightly better than random move-wise, time-wise the trained agents are much faster during evaluation because the random agent attempts to take many invalid actions before finally choosing a valid one, which increases the runtime of the evaluation.

One thing that is worth noting is that one-step curriculum leads to the least proficient agent.
The agents trained on a single level demonstrate that it is possible for the agent to almost play optimally, so this suggests that something in our way of training on a curriculum -- randomly sampling levels after an epoch -- may prevent the agent from becoming more proficient.
Improving this could be done by developing a more intelligent curriculum well as adding changes to the RL algorithm to ensure that no catastrophic forgetting will occur, where the agent forgets how to play previously learned levels.

That being said, a high proficiency does not necessarily mean that the performance of the agent is correlated with the player completion rate.
Indeed, looking at Fig. \ref{fig:correlation vs completion threshold} it can be seen that the one-step curriculum approach shows the highest correlation, despite being the least proficient agent.
It can also be seen that the highest correlation occurs when only considering $\sim$5\% of the best runs.
With the one-step curriculum scenario both being the most practical approach, due to not spending any time on additional training, and also showing the highest correlation, this is a very promising result towards using this approach in a production setting.

The least correlated approach is the two-step training, which shows an even worse correlation than random.
This might be due to the fact that it is able to completely memorise some levels, while on other levels the agent is still learning.
This mix of memorisation and proficiency may then lead to very uncorrelated behaviours.
This could also explain why in the one-step training on target levels still shows a correlation; the agent in this scenario has simply not trained long enough on a single level to memorise it, so only the agent proficiency matters.

Why the correlation with player completion rate is highest when only considering the 5\% best runs is not immediately clear but may be linked with the long tail of the move distribution: the longer the game goes on for, the more spread out the point of completion is due to an inherent randomness in the levels, leading to a lower correlation.
Conversely, there is a certain minimum number of moves required to finish many levels, so having a good run and finishing early leads to a much tighter distribution.

The results so far suggest that there is a correlation between the agent behaviour with player completion rate.
Unlike other works that try to model and predict the precise player metric (\emph{c.f.}, \cite{Gudmundsson2017}) using the rank correlation can instead be used to give an estimate on how a level is compared to other levels.
For example, it may show that a certain level is one of the top 10\% most difficult levels.

These initial results are promising but also have some limitations.
Only 120 levels were included in this analysis, which contain around 60\% of the game mechanics.
Whether these correlations extend to the remaining mechanics and whether the agent is able to deal with them need to be further investigated before using it in a production setting.
Additionally, we only consider the move distribution for the correlation.
However, it may be possible to utilise additional agent or level data for our estimates.
Not only could this possibly lead to a more robust estimate, but it could also help the level designer understand the effects of changing various aspects of a level.

\section{Conclusion}

We have examined a number of scenarios in which an RL agent can be trained and used to predict the level difficulty in a mobile puzzle game. 
The results -- based on $\sim$60\% of the game mechanics -- demonstrate that the two-step training scenario leads to the most proficient agent, while with the one-step curriculum the agent attains the largest correlation to real players' completion rates. The latter scenario is also arguably the most practical one in a production scenario.

By considering the best $\sim$5\% of the runs of the agent and record the max number of moves required to finish the level, the difficulty of the level, as measured by the player completion rate, can be estimated in terms of how it ranks compared to other levels.

Because the results shown in this research work are only for a limited subset of levels, future work should look into whether this correlation holds for the remaining levels and possibly attempt a more modelling-based approach.



\bibliographystyle{plain}
\bibliography{references,referencesmanual}

\end{document}